\documentclass{opendatalab}

\usepackage[utf8]{inputenc}
\usepackage{xcolor}
\usepackage[most]{tcolorbox}
\usepackage{longtable}
\usepackage{listings}
\usepackage[numbers]{natbib}
\usepackage{enumitem}
\usepackage{graphicx}
\usepackage{xspace} 
\usepackage{hyperref}
\usepackage{amssymb}

\usepackage{inconsolata}
\newtheorem{definition}{Definition}
\usepackage{booktabs}  % 提供更美观的三线表
\usepackage{array}     % 增强列控制
\usepackage{threeparttable} % 可选：用于带注释的表格
\usepackage{caption}     % 表格标题优化
\usepackage{enumitem}
\usepackage{colortbl} % 用于行背景色
\usepackage{cuted}  % 提供 strip 环境
\usepackage{multirow}
\usepackage{wrapfig}
\usepackage{tcolorbox}
\definecolor{iris}{HTML}{ffffff}
\definecolor{white}{HTML}{dadde2}
\definecolor{chisa}{HTML}{866f83}

\definecolor{rank1-blue}{RGB}{232, 244, 250}
\definecolor{rank2-green}{RGB}{237, 249, 237}

\newcommand{\rankone}[1]{\cellcolor{rank1-blue}\textbf{#1}}
\newcommand{\ranktwo}[1]{\cellcolor{rank2-green}#1}
\definecolor{codegreen}{rgb}{0,0.6,0}
\definecolor{codegray}{rgb}{0.5,0.5,0.5}
\definecolor{codepurple}{rgb}{0.58,0,0.82}
\definecolor{backcolour}{rgb}{0.95,0.95,0.92}
\definecolor{promptcolor}{HTML}{D1D0F2}
\definecolor{promptcolorheader}{HTML}{bdbcec}

\newcommand{\github}{\raisebox{-1.5pt}{\includegraphics[height=1.05em]{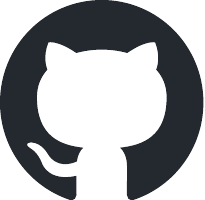}}\xspace}
\newcommand{\web}{\raisebox{-1.5pt}{\includegraphics[height=1.05em]{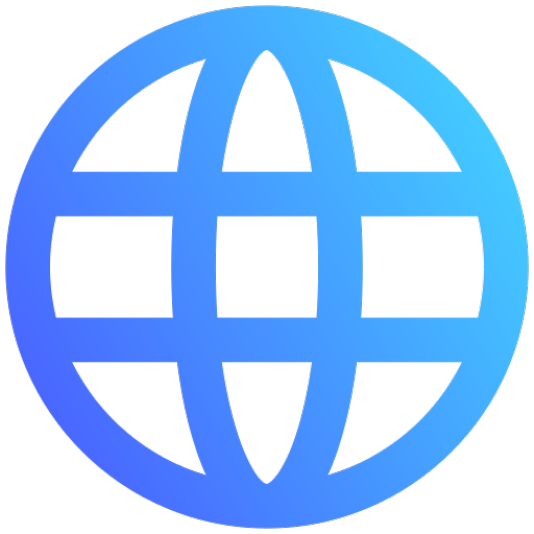}}\xspace}
\newcommand{\huggingface}{\raisebox{-1.5pt}{\includegraphics[height=1.05em]{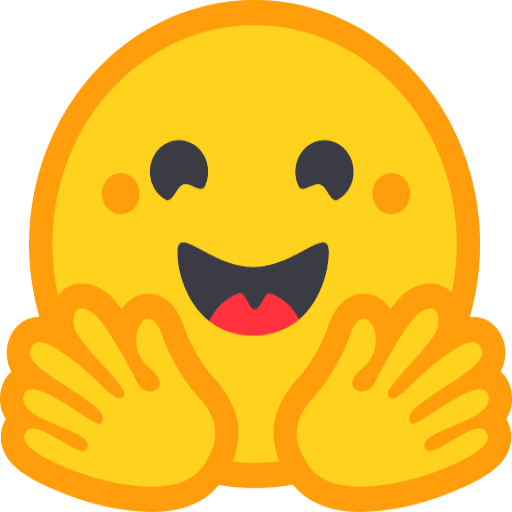}}\xspace}

\definecolor{promptcolor}{HTML}{E3F0FA}
\definecolor{promptcolorheader}{HTML}{B5D6ED}
\definecolor{prompttitletext}{HTML}{1B3A5C}

\newtcolorbox{promptbox}[1][]{
  enhanced, breakable,
  top=0.3em,bottom=0.3em,left=0.5em,right=0.5em,
  toptitle=0.3em,bottomtitle=0.2em,boxsep=0pt,
  colframe=promptcolorheader, colback=promptcolor!50, boxrule=0.5pt,
  width=\columnwidth, 
  coltitle=prompttitletext,
  title={\footnotesize #1} 
}
\lstdefinestyle{promptstyle}{
    backgroundcolor=\color{backcolour},   
    commentstyle=\color{codegreen},
    keywordstyle=\color{magenta},
    numberstyle=\tiny\color{codegray},
    stringstyle=\color{codepurple},
    basicstyle=\ttfamily\footnotesize,
    breakatwhitespace=false,         
    breaklines=true,                 
    captionpos=b,                    
    keepspaces=true,                 
    numbers=left,                    
    numbersep=5pt,                  
    showspaces=false,                
    showstringspaces=false,
    showtabs=false,                  
    tabsize=2
}
\title{Bridging Modal Isolation in Interleaved Thinking: Supervising Modality Transitions via Stepwise Reinforcement}

\author[1,2]{Tingyu Li}
\author[2]{Le Zhou}
\author[3]{Siyuan Li}
\author[1]{Yujun Wu}
\author[4]{Xinglong Xu}
\author[4]{Jingxuan Wei}
\author[1]{Conghui He}
\author[1]{Cheng Tan}

\affiliation[1]{Shanghai Artificial Intelligence Laboratory}
\affiliation[2]{Shanghai Jiaotong University}
\affiliation[3]{Zhejiang University}
\affiliation[4]{University of Chinese Academy of Sciences}

\abstract{
Interleaved thinking, where a unified multimodal model alternates between textual reasoning and visual generation, has shown promise on spatial and physical tasks. However, in complex long-chain scenarios, we identify a fundamental failure mode: generated images diverge from the textual context while subsequent text ignores the visual evidence, causing the two modalities to alternate without genuinely informing each other. We term this \textit{Modal Isolation} and attribute it to compounding information loss at modality boundaries. We decompose each reasoning cycle into atomic operations and define modality transition loss, quantifying cross-modal hallucination (text-to-image) and visual utilization deficit (image-to-text) at each boundary. We propose \textbf{MoTiF} (\textbf{Mo}dality \textbf{Ti}ransition \textbf{F}idelity), a two-stage training framework that directly optimizes these transitions: Reflective SFT trains the model to detect and recover from erroneous visual outputs; Flow-GRPO improves image generation fidelity via reinforcement learning. All training signals in MoTiF derive from transition-level fidelity rather than end-task accuracy. Across four visual puzzle benchmarks, this transition-level supervision substantially improves both cross-modal coherence and final task accuracy. The results demonstrate that effective interleaved reasoning requires explicit structural supervision at modality boundaries, not merely scaling or end-task optimization.
}

\metadatainfo{
    \small\makebox[\linewidth][c]{
        \github~\href{https://github.com/OpenRaiser/MoTiF}{\textbf{Code}} \quad
        \web~\href{https://openraiser.github.io/MoTiF/}{\textbf{Website}} \quad
        \huggingface~\href{https://huggingface.co/datasets/OpenRaiser/MoTiF}{\textbf{Dataset}}
    }
}

\begin{document}

\maketitle

% \tableofcontents

\section{Introduction}

The ability to reason about the physical world requires coordinating abstract symbolic thinking with concrete visual imagination~\cite{battaglia2013simulation,yu2026latent,wei2026trinity}. Unified multimodal models (UMMs), capable of natively generating both text and images~\cite{team2024chameleon,xie2025show}, have enabled \textit{interleaved thinking}: a reasoning paradigm where the model alternates between textual reasoning and visual world modeling within a single chain~\cite{li2025imagine,chern2025thinking,gu2025thinkmorph, visworld}. Prior work has demonstrated the viability of this paradigm on short-horizon spatial and physical tasks. In this work, we investigate its behavior in long-chain visual reasoning.

\begin{wrapfigure}{r}{0.4\linewidth}   % {r}表示图片靠右，文字在左侧；想靠左则用{l}
    \centering
    \includegraphics[width=\linewidth]{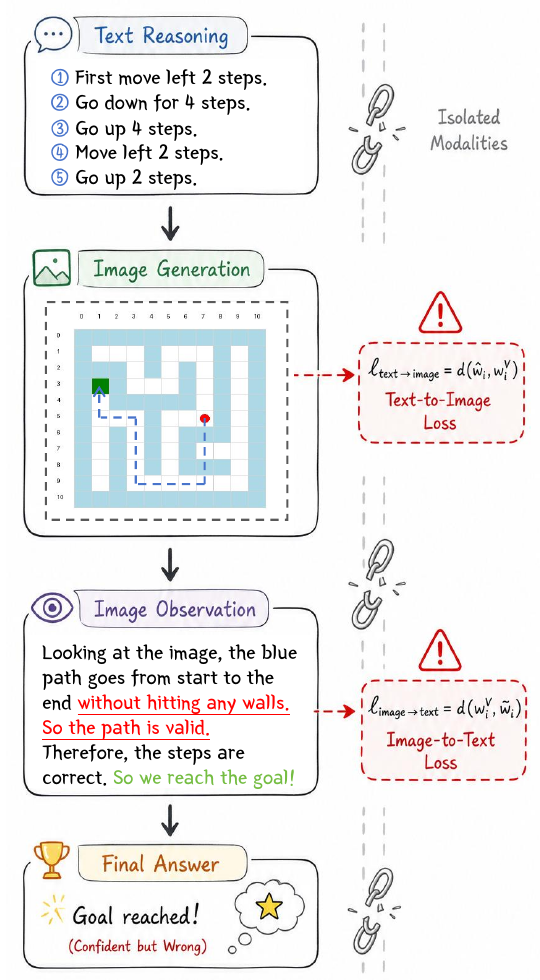}
    \vspace{-7mm}
    \caption{The phenomenon of modal isolation in interleaved thinking. In a maze navigation task, the generated image depicts an inconsistent path. Subsequently, the model observes the erroneous image yet fails to detect the discrepancy, incorrectly validating the path.}
    \label{fig:isolation}
    \vspace{-20mm}
\end{wrapfigure}

We observe a persistent failure mode across multiple tasks, illustrated in Figure~\ref{fig:isolation}. Consider a maze navigation example: the model first performs textual reasoning, planning a sequence of moves (``move left 2 steps, go down 4 steps, go up 4 steps...''). It then generates an image to visualize the resulting path. However, the generated image depicts a path that is fundamentally inconsistent with the textual plan—passing through walls and deviating from the specified moves. This constitutes \textit{text-to-image} information loss ($\ell_{\text{text} \to \text{image}}$). More critically, when the model subsequently observes its own generated image, it fails to detect these inconsistencies: it declares ``the blue path goes from start to end \textit{without hitting any walls}'' and confidently concludes the goal is reached—despite the path being clearly invalid. This constitutes \textit{image-to-text} information loss ($\ell_{\text{image} \to \text{text}}$). The two modalities alternate in form but never genuinely inform each other, and the model arrives at a \textit{confident but wrong} answer. We term this phenomenon \textbf{Modal Isolation}: compounding information loss at modality boundaries progressively decouples the two modalities, causing each to operate in isolation despite their surface-level interleaving.

A natural response is to optimize end-task accuracy via reinforcement learning, but this approach is fundamentally misaligned with the problem: it can reward models that arrive at correct answers 
through hallucinated intermediate reasoning, leaving modal isolation unaddressed. Similarly, supervised fine-tuning on correct chains teaches the model \textit{what} to produce but not \textit{how to verify} cross-modal consistency, offering no mechanism to detect or recover from generation errors at inference time. These limitations point to the need for \textit{process-level supervision at modality boundaries}.

We formalize the information loss at each modality boundary as \textit{Modality Transition Loss}, decomposing it into cross-modal hallucination (text$\to$image fidelity) and visual utilization deficit (image$\to$text fidelity). We then propose \textbf{MoTiF} (\textbf{Mo}dality \textbf{Ti}ransition \textbf{F}idelity), a two-stage training framework that directly optimizes a joint transition objective, with neither stage relying on end-task accuracy: (1) \textit{Reflective SFT}, which trains the textual modality to detect and recover from erroneous visual outputs by exposing the model to deliberately corrupted images during training; and (2) \textit{Flow-GRPO}, which improves the visual modality's fidelity to textual instructions via reinforcement learning over the image generation process.

Our main contributions are: 
\begin{itemize}[leftmargin=2em]
\item We identify and formalize modal isolation, a compounding failure mode in interleaved thinking where modalities alternate without mutual information transfer. 
\item We propose a two-stage training pipeline that provides structural supervision at modality boundaries without relying on end-task accuracy signals. 
\item Across four visual puzzle benchmarks, our method significantly improves the performance, demonstrating that transition-level supervision is both necessary and effective for long-chain interleaved reasoning. 
\end{itemize}

\section{Related Work}

\paragraph{Multimodal Understanding and Generation.}
% 近年来，多模态模型逐渐从以感知和理解为中心的任务，扩展到理解与生成的联合建模，推动了统一多模态模型（UMMs）的发展。早期系统通常将大语言模型与图像生成器、视觉解码器或生成式视觉 token 连接起来，使模型能够在多模态上下文条件下生成图像或混合模态输出。后续工作进一步探索语言与视觉之间更紧密的统一建模方式，包括基于多模态 token 的自回归建模、早期融合的混合模态架构、统一的自回归-扩散框架，以及原生图文生成。近期工作则进一步审视统一框架中视觉理解与图像生成之间的关系，指出二者可能需要不同粒度的视觉表征，并通过视觉编码解耦、模型扩展和大规模统一预训练提升理解与生成的协同。这些研究建立了文本与图像能够在共享多模态接口中被共同建模和生成的条件。然而，它们主要关注如何构建能够跨模态理解与生成的模型，而不是这些能力在长链推理中被反复组合时会呈现出怎样的动态行为。
Recent multimodal models have gradually moved beyond perception-centered understanding toward joint modeling of understanding and generation, contributing to the development of unified multimodal models (UMMs). Early approaches connect large language models with image generators, visual decoders, or generative visual tokens, enabling models to produce images or mixed-modal outputs conditioned on multimodal contexts~\cite{koh2023generating,dong2024dreamllm,minigpt5_2023,sun2024emu,sun2024generative}. Subsequent work explores tighter integration between language and vision, including autoregressive modeling over multimodal tokens, early-fusion mixed-modal architectures, unified autoregressive-diffusion frameworks, and native image-text generation~\cite{wang2026multimodal,chameleon2024,xie2025showo,anole2024}. More recent studies further revisit the relationship between visual understanding and image generation within unified frameworks, showing that the two capabilities may require visual representations at different levels of granularity, and improving their coordination through decoupled visual encoding, model scaling, and large-scale unified pretraining~\cite{wu2025janus,januspro2025,bagel2025}. Together, these studies establish the conditions under which text and images can be jointly modeled and generated through a shared multimodal interface. However, they primarily address how to build models capable of understanding and generation across modalities, rather than how these capabilities behave when repeatedly composed over long reasoning.
% chains.
\paragraph{Interleaved Visual Reasoning.}
% 在上述模型能力的基础上，interleaved visual reasoning 进一步研究视觉表征如何参与推理过程本身，而不只是作为静态输入或最终输出。显式视觉状态生成方法通过生成视觉想象、视觉草图或渐进式图文推理轨迹，辅助空间关系、物体状态变化和多步视觉规划等任务。视觉证据锚定的多模态推理方法则将 grounding 信号、视觉 token 或结构化图文监督引入 Chain-of-Thought，在不一定合成完整图像的情况下增强语言推理与视觉证据之间的关联。后训练方法进一步通过强化学习或过程级奖励鼓励模型获取、生成或利用视觉信息，从而提升复杂推理中对中间视觉状态的建模和使用能力。相关评测与分析表明，中间视觉信息能够提供纯语言推理难以显式表达的状态约束，尤其适用于空间结构、物理状态追踪和视觉世界建模任务。总体来看，这些研究确立了视觉内容作为多模态推理中间表征的作用。然而，现有工作主要考察视觉步骤的有效性、构造方式、激发策略及其对最终任务表现的影响；相较之下，模态转换本身的保真性仍缺乏显式刻画。具体而言，文本推理中编码的世界状态可能未能被生成图像忠实外化，后续文本也可能未能正确解码或利用已经生成的视觉证据。这一缺口构成了本文研究 Modal Isolation 的直接动机，即长链推理中两种模态在形式上交替推进，却在世界状态传递上逐步脱耦。
Building on these model-level advances, interleaved visual reasoning studies how visual representations can participate in the reasoning process itself, rather than serving only as static inputs or final outputs. Explicit visual-state generation methods produce visual imagination, sketches, or progressive image-text reasoning traces to support tasks involving spatial relations, object-state changes, and multi-step visual planning~\cite{li2025imagine,twgi2025,gu2025thinkmorph}. Visually grounded multimodal reasoning methods instead introduce grounding signals, visual tokens, or structured image-text supervision into chain-of-thought reasoning, strengthening the connection between verbal reasoning and visual evidence without necessarily synthesizing full images~\cite{fan2026grit,gao2025interleaved,chen2026mint,qin2025unicot}. Post-training approaches further encourage models to acquire, generate, or exploit visual information through reinforcement learning or process-level rewards, improving the modeling and use of intermediate visual states in complex reasoning~\cite{zheng2025deepeyes,nie2026towards}. Complementary benchmarks and analyses show that intermediate visual information can provide state constraints that are difficult to express through language alone, especially in tasks involving spatial structure, physical state tracking, and visual world modeling~\cite{zhou2025visualizing,visworld}. Taken together, these studies establish visual content as an intermediate representation for multimodal reasoning. However, existing work primarily examines the utility, construction, elicitation, and final-task benefits of visual steps; the fidelity of the modality transitions themselves remains less explicitly characterized. Specifically, the world state encoded in textual reasoning may fail to be faithfully externalized by the generated image, and subsequent text may fail to correctly decode or use the generated visual evidence. This gap motivates our focus on Modal Isolation, where two modalities appear to alternate over long reasoning chains but progressively decouple in the world states they encode and use.

\section{Method}

\subsection{Preliminaries: Interleaving Thinking}

Let $\mathcal{W}$ be the world state space of a visual reasoning task—encoding the positions, spatial relations, and physical configurations of entities in the environment. Let $d: \mathcal{W} \times \mathcal{W} \to \mathbb{R}_{\geq 0}$ be a divergence measure on $\mathcal{W}$ that quantifies the discrepancy between two world states.\\
An interleaving thinking chain over $K$ visual generation steps is a sequence
\begin{equation}
\mathbf{R} = (t_1, v_1, t_2, v_2, \ldots, v_K, t_{K+1}),
\label{eq:chain}
\end{equation}
where $t_i \in \mathcal{T}$ denotes a text segment, $v_i \in \mathcal{V}$ a generated image, and each element is conditioned on all preceding elements in the chain.

Rather than treating $t_i$ as an opaque unit, we decompose it by function within the reasoning cycle. For each step $i$, the text segment $t_i$ performs three atomic operations:
\begin{enumerate}
    \item \label{eq:observe}
          \(t_i^{\text{obs}}\) extracts a world state estimate \(\tilde{w}_{i-1} \in \mathcal{W}\) from the preceding image \(v_{i-1}\);
    \item \label{eq:reason}
          \(t_i^{\text{rea}}\) performs symbolic reasoning over \(\tilde{w}_{i-1}\) to decide the next action;
    \item \label{eq:instruct}
          \(t_i^{\text{ins}}\) encodes the intended next world state \(\hat{w}_i \in \mathcal{W}\) as a drawing instruction.
\end{enumerate}
The image $v_i$ is then generated conditioned on $t_i$, completing the cycle. We write $t_i = (t_i^{\text{obs}}, t_i^{\text{rea}}, t_i^{\text{ins}})$ to denote this functional decomposition.

Two modality transitions occur in each cycle, at the boundaries between textual and visual processing:
\begin{align}
\mathcal{M}_{\mathcal{T} \to \mathcal{V}}^{(i)} &: t_i^{\text{ins}} \mapsto v_i, \label{eq:M_und_gen} \\
\mathcal{M}_{\mathcal{V} \to \mathcal{T}}^{(i)} &: v_i \mapsto t_{i+1}^{\text{obs}}. \label{eq:M_gen_und}
\end{align}

\begin{figure}[t]
\centering
\includegraphics[width=1.0\linewidth]{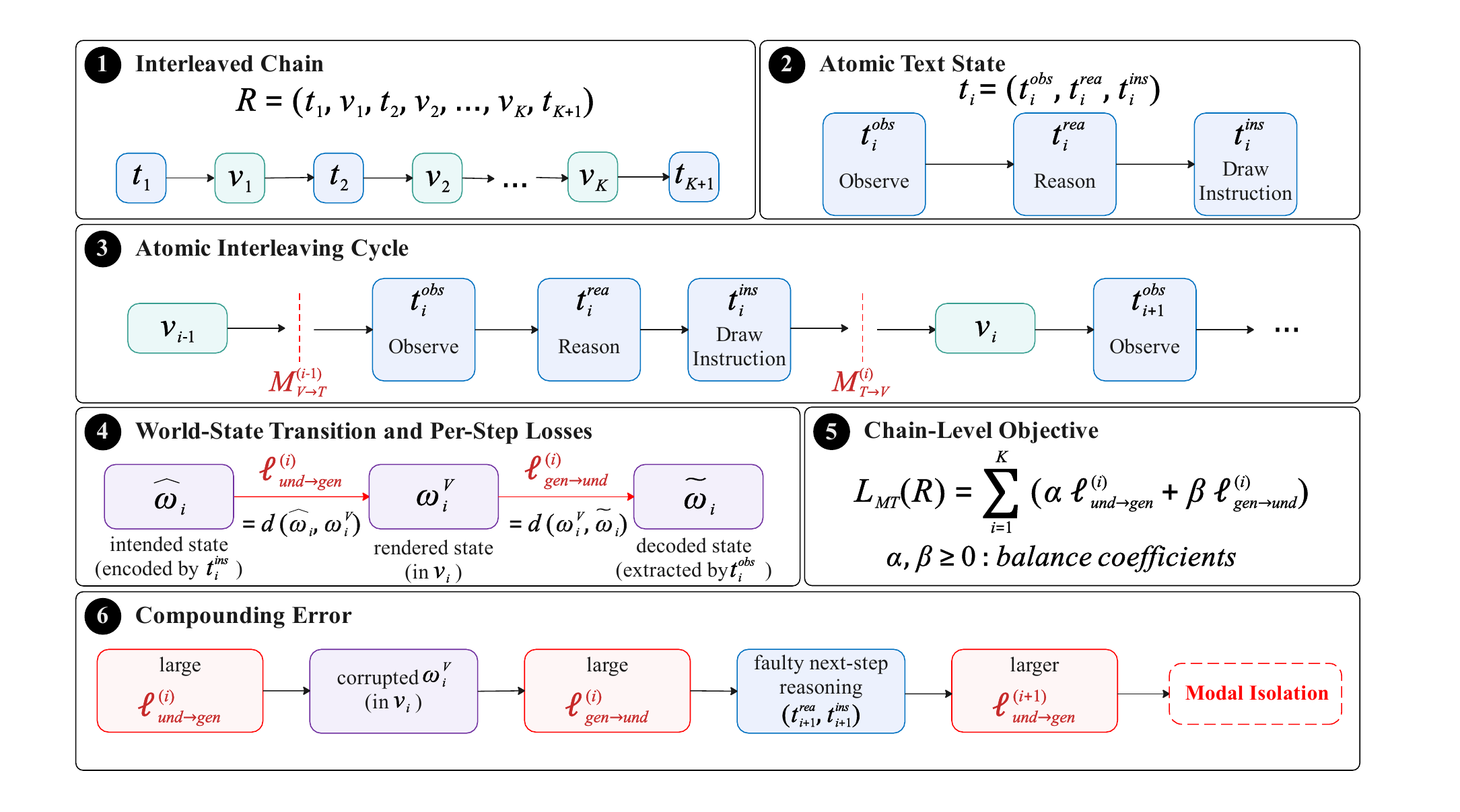} % or .png/.jpg
% \vspace{-10mm}
\caption{Modality transition loss and atomic decomposition of interleaving thinking.}
% \vspace{-5mm}
\label{fig:intro}
\end{figure}

At each transition, information crosses between two heterogeneous representation spaces: the symbolic space $\mathcal{T}$ (discrete, compositional, structured by linguistic syntax and semantics) and the pixel space $\mathcal{V}$ (continuous, holistic, structured by visual statistics). These spaces encode world states through fundamentally different inductive biases. Every modality crossing is therefore a potential site of information loss—the encoding modality must faithfully externalize its world knowledge, and the decoding modality must correctly interpret what was externalized.

\subsection{Modality Transition Optimization}

\textbf{Modality Transition Loss.} To quantify the information loss at modality boundaries, we introduce three world state variables for each cycle $i$:
\begin{itemize}
    \item $\hat{w}_i \in \mathcal{W}$: the world state that $t_i^{\text{ins}}$ \textit{intends} to render—the target state produced by symbolic reasoning.
    \item $w_i^V \in \mathcal{W}$: the world state \textit{actually rendered} in $v_i$—what the visual modality produced in pixel space.
    \item $\tilde{w}_i \in \mathcal{W}$: the world state that $t_{i+1}^{\text{obs}}$ \textit{extracts} from $v_i$—what the symbolic modality decodes from visual evidence.
\end{itemize}
These three variables trace the lifecycle of a world state as it traverses a modality boundary: encoding ($\hat{w}_i$), rendering ($w_i^V$), and decoding ($\tilde{w}_i$). When the transition is lossless, $\hat{w}_i = w_i^V = \tilde{w}_i$; in practice, each step introduces distortion.

\begin{definition}[Cross-Modal Hallucination]
For the und → gen transition $\mathcal{M}_{\mathcal{T} \to \mathcal{V}}^{(i)}$, the cross-modal hallucination loss is
\begin{equation}
\ell_{\text{und}\to\text{gen}}^{(i)} \triangleq d(\hat{w}_i, w_i^V).
\label{eq:cmh}
\end{equation}
This measures the degree to which $v_i$ diverges from the intended world state $\hat{w}_i$: a large value indicates that the visual modality failed to faithfully externalize the reasoning output of the textual modality.
\end{definition}

\begin{definition}[Visual Utilization Deficit]
For the gen → und transition $\mathcal{M}_{\mathcal{V} \to \mathcal{T}}^{(i)}$, the visual utilization deficit is
\begin{equation}
\ell_{\text{gen}\to\text{und}}^{(i)} \triangleq d(w_i^V, \tilde{w}_i).
\label{eq:vud}
\end{equation}
This measures the degree to which the text modality fails to correctly decode the visual evidence in $v_i$: a large value indicates that textual reasoning proceeds on an incorrect or incomplete world model despite the availability of visual information.
\end{definition}

Summing over the chain yields the \textbf{Modality Transition Loss}:
\begin{equation}
\mathcal{L}_{\text{MT}}(\mathbf{R}) = \sum_{i=1}^{K} \left( \alpha \cdot \ell_{\text{und}\to\text{gen}}^{(i)} + \beta \cdot \ell_{\text{gen}\to\text{und}}^{(i)} \right),
\label{eq:mt_loss}
\end{equation}
where $\alpha, \beta > 0$ are weighting coefficients. This definition is self-contained at the conceptual level: it depends only on the abstract structure of $\mathcal{W}$ and $d$, without reference to how $d(\cdot, \cdot)$ is empirically estimated.\footnote{In our experiments, $d(\cdot, \cdot)$ is operationalized through rubric-based VLM-as-Judge evaluations; see Section~\ref{sec:exp} for details. The theoretical framework is independent of this implementation choice.}

The two components of $\mathcal{L}_{\text{MT}}$ are not independent; their interaction produces a compounding effect across the chain. When $\ell_{\text{und}\to\text{gen}}^{(i)}$ is large, $w_i^V$ is a degraded representation of the true world state, which raises the expected value of $\ell_{\text{gen}\to\text{und}}^{(i)}$—the text must now decode a corrupted visual signal. When $\ell_{\text{gen}\to\text{und}}^{(i)}$ is large, $\tilde{w}_i$ (and consequently $\hat{w}_{i+1}$ encoded by $t_{i+1}^{\text{ins}}$) rests on a faulty perceptual foundation, making $\ell_{\text{und}\to\text{gen}}^{(i+1)}$ more likely to be large. This mutual amplification accumulates with chain length $K$, providing a formal account of \textbf{Modal Isolation}: the progressive decoupling of textual reasoning from visual evidence, where each modality drifts further from the other with every transition. $\mathcal{L}_{\text{MT}}$ thus serves as both a diagnostic metric and a fundamental bottleneck on the scalability of interleaving thinking to long chains.

\textbf{Joint Optimization of Modality Transitions.} The above diagnosis suggests a direct remedy: rather than optimizing for end-task accuracy alone—which may inadvertently reward models that arrive at correct answers through hallucinated reasoning—we directly supervise the modality transitions themselves. The training objective is
\begin{equation}
\min_\theta \; \mathbb{E}_{\mathbf{R} \sim \pi_\theta} \left[ \mathcal{L}_{\text{MT}}(\mathbf{R}) \right].
\label{eq:mt_obj}
\end{equation}

We decompose this objective by transition direction:
\begin{align}
\mathcal{L}_{\text{und}\to\text{gen}}(\theta) &= \mathbb{E}_{\mathbf{R} \sim \pi_\theta} \left[ \sum_{i=1}^{K} \ell_{\text{und}\to\text{gen}}^{(i)} \right], \label{eq:obj_und_gen} \\
\mathcal{L}_{\text{gen}\to\text{und}}(\theta) &= \mathbb{E}_{\mathbf{R} \sim \pi_\theta} \left[ \sum_{i=1}^{K} \ell_{\text{gen}\to\text{und}}^{(i)} \right]. \label{eq:obj_gen_und}
\end{align}
The two sub-objectives govern different generation policies: $\mathcal{L}_{\text{und}\to\text{gen}}$ concerns the image generation policy $\pi_\theta^V$ (a flow matching model), while $\mathcal{L}_{\text{gen}\to\text{und}}$ concerns the text generation policy $\pi_\theta^T$ (an autoregressive model). Due to this architectural heterogeneity, we adopt a \textit{jointly optimized but separately implemented} strategy: both sub-objectives share the same base model parameters $\theta$ and jointly minimize $\mathcal{L}_{\text{MT}}$, while each is optimized through the RL framework best suited to its modality.

For $\mathcal{L}_{\text{und}\to\text{gen}}$, we formulate the flow matching denoising process as a Markov Decision Process and apply online policy optimization via Flow-GRPO~\cite{flow-grpo}. Given a ground-truth textual context ending with 
\begin{wrapfigure}{r}{0.45\linewidth}   % 靠右，宽度占当前行宽的45%
    \centering
    \includegraphics[width=\linewidth]{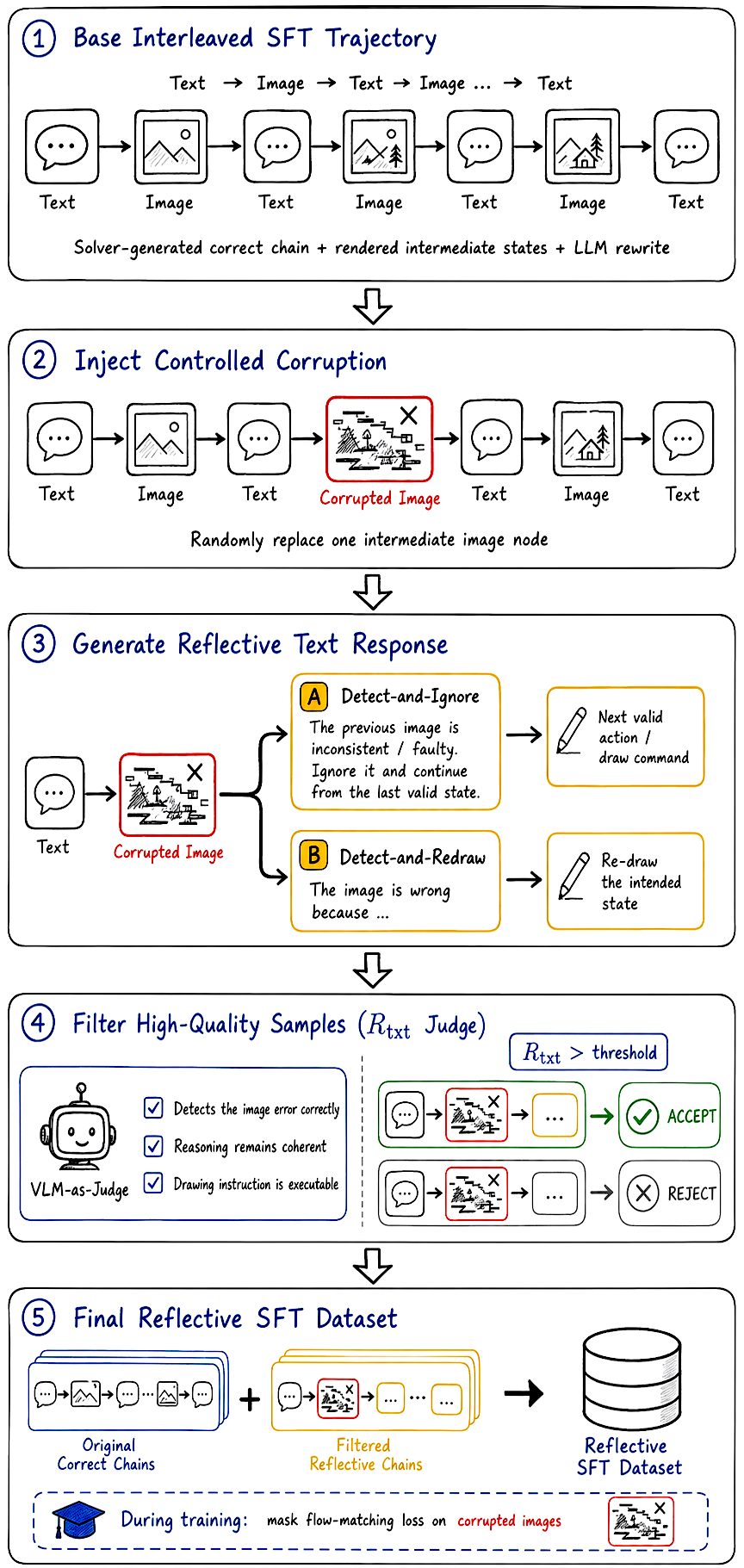}
    \caption{Reflective SFT data collection pipeline.}
    \label{fig:reflexion}
    \vspace{-48mm}
\end{wrapfigure}
$t_i^{\text{ins}}$, the model generates an image $v_i$ and receives a reward signal derived from the empirical estimate of $\ell_{\text{und}\to\text{gen}}^{(i)}$. The policy $\pi_\theta^V$ is updated via group-relative advantage estimation with KL divergence regularization against a reference policy, maximizing the fidelity of image generation to textual instructions while preserving generation quality.

For $\mathcal{L}_{\text{gen}\to\text{und}}$, we construct contrastive supervision data by introducing controlled perturbations to the visual stream: an intermediate image $v_i$ in a ground-truth chain is replaced with a corrupted variant $\tilde{v}_i$, and a frontier model is prompted to produce text that detects the discrepancy and either corrects the visual output or works around it. The resulting data, filtered by the empirical estimate of $\ell_{\text{gen}\to\text{und}}^{(i)}$, provides supervised signals that explicitly train the textual modality to attend to, verify, and recover from visual evidence—thereby reducing the visual utilization deficit.

Neither optimization path depends on end-task accuracy. Every training signal is derived from the fidelity of a single modality transition, evaluated independently of whether the final answer is correct. The two paths are unified under the single objective $\min_\theta \mathcal{L}_{\text{MT}}$: both serve to reduce information loss at modality boundaries, and both contribute to mitigating modal isolation and enabling more coherent long-chain interleaving thinking.

\begin{table}[htbp]
% 仅针对当前表格设置标题靠左
\captionsetup{justification=raggedright, singlelinecheck=false} 
\caption{Naive interleaving SFT datasets.}
\label{tab:tasks}
\small
\begin{tabular}{l l c} 
\toprule
\textbf{Task} & \textbf{Reference} & \textbf{Size} \\
\midrule
Sokoban      & Game-RL           & 7,997 \\
Maze         & Game-RL           & 7,365 \\
Multi-hop Manipulation & CLEVER  & 8,000   \\
Ball Tracking & RBench-V & 8,000   \\
\bottomrule
\end{tabular}
\end{table}

%%%%%%%%%%%%%%%%%%%%%%%%%%%%%%%%%%%%%%%%%%%%%%%%%%%%%%

\section{Experimental Design}
\label{sec:exp}
In this chapter, we systematically introduce the task selection and data construction pipeline, training workflow, hyperparameter settings, and evaluation suite.
% , thereby providing an executable recipe for subsequent work.
% 在这一章节中我们将系统性的介绍任务选择和数据构建流程，训练流程，实验参数设置以及评估套件；为之后工作的复现提供可执行的recipe

\subsection{Task Selection and Naive SFT Dataset}
Based on the varying cognitive load required to solve multimodal problems (i.e., whether the primary reasoning function is performed by the generation or the understanding module), we select four representative types of tasks~\cite{tong2025game}: \textbf{(1) Sokoban.} Built upon Game-RL. This task requires the model to find the optimal path (i.e., pushing boxes to designated targets with the minimum number of steps) in a Sokoban grid environment. In this task, the understanding module bears the primary reasoning

responsibility. \textbf{(2) Maze.} Also built upon the Game-RL codebase, this task requires the model to find the correct path from the start to the goal in a maze environment. \textbf{(3) Multi-hop Manipulation.} Inspired by works such as CLEVER, we use Blender to render composite images that contain certain spatial relationships and individual attributes. The model is required to perform operations such as “removing/adding” objects, converting object attributes (e.g., color, shape), and moving objects’ spatial positions, and then answer specific spatial understanding questions. In tasks (2) and (3), the model needs to combine the abstract symbolic reasoning provided by the visual understanding module with the physical world intuition provided by the generation module to solve the problems. \textbf{(4) Ball Tracking.} Based on RBench-V and Viseval, this task provides the starting position and outgoing direction of a ball, and asks after several specular reflections which region the ball finally reaches. This task heavily demands the model’s physical intuition and visual world knowledge modeling, and the visual generation module bears the primary reasoning responsibility.
We use a rule-based solver to obtain the correct answers and code-synthesized chain-of-thought; meanwhile, we render the key intermediate images during the solving process. We prompt Gemini3.1-Flash-Lite to rewrite the chain-of-thought, ultimately obtaining a high-quality, image-text interleaved fine-tuning dataset. See Table~\ref{tab:tasks}.

% 我们根据解决多模态问题所需的认知负载不同（即承担主要推理职能的模块是生成还是理解），选定了4类具有代表性的任务：\textbf{(1). Sokoban.} 基于 Game-RL 进行构建. 该任务要求模型在推箱子网格环境中找到最优路径，即将箱子用最少步数推到指定终点，在该任务中承担主要推理职责的是理解模块。\textbf{(2). Maze.} 同样基于 Game-RL 源码进行构建，该任务要求模型在迷宫环境中找到正确的从起点到终点的路径。\textbf{(3). Manipulation.} 我们受 CLEVER 等工作的启发，基于blender渲染具有一定空间关系，个体属性的集合体图像，要求模型执行“移除/添加” 物体，转换物体颜色/形状等个体属性，以及移动物体空间位置等操作，然后询问具体的空间理解问题。(2), (3)两类任务中模型需要结合视觉理解模块提供的抽象符号推理以及生成模态提供的物理世界直觉来解决问题。\textbf{(4). Ball Tracking.} 基于RBench-V以及Viseval，该问题给定了一个小球的起始位置以及出射方向，询问数次镜面反射之后，小球最终到达的区域序号；该问题极度要求模型的物理直觉和视觉世界知识建模，承担主要推理职责的是视觉生成模块。
% 我们基于规则求解器，得到正确的答案和基于代码合成的思维链；同时渲染得到求解过程中的关键中间图像。我们提示 Gemini3.1-Flash-Lite 重写该思维链，最终得到具有高质量的图文交错微调数据集

\subsection{Model Training}
To isolate the effects of Modal Isolation across tasks, we train a separate model per task. We choose Bagel~\cite{bagel2025} as our baseline, and all experiments are done on 8$\times$H200 GPUs, full hyperparameters are in the Appendix. Our pipeline has two stages mapping to the two sub-objectives of Eq.~\ref{eq:mt_obj}.

\noindent\textbf{Reward Design.}
To operationalize the modality transition losses, we construct two rubric-based VLM-as-Judge evaluators that explicitly quantify cross-modal hallucination and visual utilization at each modality boundary. Both produce binary (0/1) judgments, providing clean, transition-level training signals independent of end-task outcomes. Concretely, $R_{\text{img}}$ evaluates the $\mathcal{M}_{\mathcal{T} \to \mathcal{V}}$ transition: given the problem, the preceding reasoning context ending with a drawing instruction, the generated image, and the ground-truth image, it judges whether the generated image is \textit{logically equivalent} to the ground truth---requiring agreement on task-relevant objects, spatial relations, and actions while tolerating superficial style differences. $R_{\text{txt}}$ evaluates the $\mathcal{M}_{\mathcal{V} \to \mathcal{T}}$ transition: for non-final steps it checks whether the \texttt{<think>...</think>} text correctly observes the preceding image, reasons coherently, and issues an executable drawing instruction; for final steps it verifies answer correctness and \texttt{<ans>...</ans>} formatting. Both evaluators are implemented by prompting a strong frontier VLM with detailed rubrics.
% (see Appendix~\ref{app:prompts} for the full prompt design).

\noindent\textbf{Stage 1: Reflective SFT.}
This stage optimizes $\mathcal{L}_{\text{gen}\to\text{und}}$ by training the model to detect and recover from erroneous visual outputs. Starting from the naive interleaving chains in Section~\ref{sec:exp}, we randomly replace an intermediate image $v_k$ with a corrupted variant $\tilde{v}_k$ to simulate a generation error, then prompt Gemini3.1-Flash-Lite to produce reflective text that identifies the discrepancy and responds. Two recovery patterns are used:\textbf{(1) Detect-and-Ignore.} where the model acknowledges the error and continues without regenerating. \textbf{(2) Detect-and-Redraw.} where it diagnoses the error and re-issues the drawing instruction for $v_k$. The resulting chains are filtered by $R_{\text{txt}}$, retaining only samples where the reflection correctly addresses the perturbation. 
% This yields a dataset that explicitly supervises the $\mathcal{M}_{\mathcal{V} \to \mathcal{T}}$ transition.

We fine-tune Bagel with a combined objective:
\begin{equation}
\mathcal{L}_{\text{SFT}}(\theta) = -\sum_{i} \log p_\theta(t_i \mid C_{<i}) + \sum_{j \notin \mathcal{M}} \mathbb{E}_{t, \epsilon} \left\| v_\theta(x_j^t, t \mid C_{<j}) - (\epsilon - x_j) \right\|_2^2
\label{eq:sft_loss}
\end{equation}
where $\mathcal{M}$ indexes the perturbed images $\tilde{v}_k$, whose flow matching loss is masked to prevent the model from learning erroneous visual states. The model is trained to follow the format \texttt{<think>...</think><create\_image>} for intermediate steps and \texttt{<ans>...</ans>} for the answer.

\noindent\textbf{Stage 2: Flow-GRPO.}
This stage optimizes $\mathcal{L}_{\text{und}\to\text{gen}}$ by improving the fidelity of image generation to textual drawing instructions. We extracted prefixes $[Q, t_1, v_1, \ldots, t_i]$ from ground-truth chains, where $t_i$ ends with a drawing instruction $t_i^{\text{ins}}$; the model must generate $v_i$. 

Following Flow-GRPO~\cite{flow-grpo}, we formulate the flow matching denoising process as an MDP and optimize via group-relative advantage estimation, using $R_{\text{img}}$ as the binary reward at the final denoising step:
\begin{equation}
\mathcal{L}_{\text{Flow-GRPO}} = -\log \pi_{\theta} \cdot \hat{A} + \beta\, D_{\mathrm{KL}}(\pi_{\theta} \| \pi_{\mathrm{ref}}),
\label{eq:flow_grpo_core}
\end{equation}
We adopt the policy gradient formulation for concise presentation. Refer to the appendix for details.

%%%%%%%%%%%%%%%%%%%%%%%%%%%%%%%%%%%%%%%%
\begin{table*}[t]
\small
\centering
\caption{Model performance on our benchmarks.}
\label{tab:model_performance}
    \resizebox{1.0\linewidth}{!}{
\begin{tabular*}{\textwidth}{@{\extracolsep{\fill}}l|cccc|c@{}}
\toprule
\textbf{Model} 
& \multicolumn{1}{c}{\textbf{Sokoban}} 
& \multicolumn{1}{c}{\textbf{Maze}} 
& \multicolumn{1}{c}{\textbf{Manipulation}} 
& \multicolumn{1}{c|}{\textbf{Ball Tracking}} 
& \multicolumn{1}{c}{\textbf{Overall}} \\
\midrule
%\rowcolor{green!10} 
\multicolumn{6}{l}{\textit{Frontier Models}} \\
\midrule
Gemini3.5-Flash      & 85.71 & 97.47 & 93.33 & 77.33 & 88.46 \\
Gemini3.1-Flash-Lite & 37.30 & 11.39 & 88.00 & 40.00 & 44.17 \\
Seed2.0-Lite-0215    & 69.05 & 64.56 & 89.33 & 58.67 & 70.40 \\
\midrule
%\rowcolor{green!10} 
\multicolumn{6}{l}{\textit{Open-Source Models}} \\
\midrule
Qwen3.5-27B \cite{qwen3.5}         & \rankone{80.95} & 16.46 & 86.67 & 32.00 & 54.02 \\
Gemma-4-31B         & 66.67 & 13.92 & \ranktwo{90.67} & 32.00 & 50.82 \\
Bagel-7B-MoT \cite{bagel2025}        & 16.67 & 10.13 & 45.33 & 29.33 & 25.37 \\
ThinkMorph \cite{gu2025thinkmorph} & 18.25 & 8.86 & 69.33 & 48.00 & 36.11 \\
\quad $\vdash$ w.o. interleave thinking & 22.22 & 5.06 & 65.33 & 37.33 & 31.74 \\
\midrule
%\rowcolor{green!10} 
\multicolumn{6}{l}{\textit{Our Models}} \\
\midrule
Ours (Stage1: optimize. $\mathcal{L}_{\text{gen}\to\text{und}}$) & 43.65 & \ranktwo{67.09} & 87.67 & \ranktwo{70.33} & \ranktwo{67.19} \\
Ours (Stage2: optimize. $\mathcal{L}_{\text{und}\to\text{gen}}$) & \ranktwo{50.00} & \rankone{70.57} & \rankone{90.67} & \rankone{72.67} & \rankone{70.98} \\
% \midrule
$\Delta$ max (vs. BaseModel) & $+33.33$ & $+60.44$ & $+45.34$ & $+43.34$ & $+45.61$ \\
\bottomrule
\end{tabular*}
}
\end{table*}

\begin{figure}[ht]
\centering
% \vspace{-4mm}
\includegraphics[width=\linewidth]{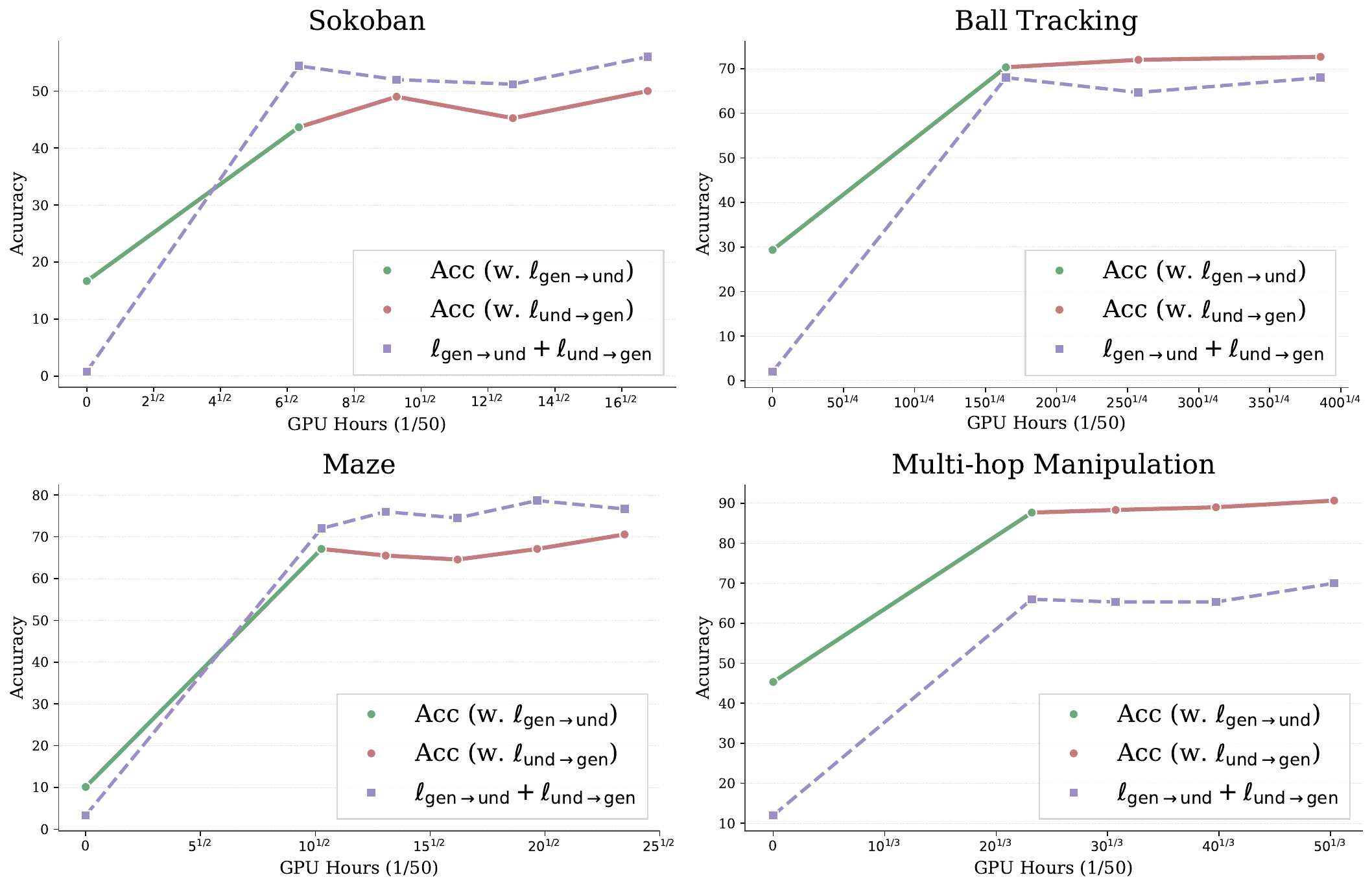} % or .png/.jpg
\caption{Solid lines denote model accuracy scores, with different colors distinguishing the Reflective SFT and Flow-GRPO training stages. Dashed lines represent the model's cross-modal reward scores. This figure illustrates how model capabilities evolve synergistically with the optimization of Modal Transition Loss across various tasks.} % 实线表示模型准确率得分，用不同颜色区分了Reflective SFT和Flow-GRPO两段训练。虚线代表模型的跨模态奖励得分。该图展示了在不同tasks，模型能力是如何随着Modal Transition Loss的优化而协同进化的
% \vspace{-4mm}
\label{fig:acc}
\end{figure}

\section{Results Analysis}
We demonstrate that even without explicit outcome reward supervision, leveraging stepwise rewards based on VLM-as-Judge to mitigate modal isolation in interleaved thinking helps boost the model's performance on visual reasoning tasks. Additionally, we discuss whether Reflective SFT and Flow-GRPO in the model training pipeline—whose objectives are to optimize $\mathcal{L}_{\text{und}\to\text{gen}}$ and $\mathcal{L}_{\text{gen}\to\text{und}}$ respectively—incur catastrophic forgetting and damage the model's learned capabilities.
% 在这一个章节，我们展示了，即便不进行显式的结果奖励监督，通过基于VLM-as-Judge的stepwise reward，缓解Interleaving Thinking存在的modal isolation现象，可以有助于提升模型解决视觉推理问题的能力(section 5.1)。同时，我们讨论了模型训练pipeline中，Reflective SFT和Flow-GRPO（目标分别是优化$\mathcal{L}_{\text{und}\to\text{gen}}$，$\mathcal{L}_{\text{gen}\to\text{und}}$）是否会存在灾难性遗忘，损害模型已习得能力

\subsection{Modality Transition Optimization Boosts Interleaving Thinking}
Table~\ref{tab:model_performance} presents the performance of frontier multimodal models and open-source multimodal reasoning models with a similar parameter scale to Bagel-7B-MoT across four tasks (Sokoban, Maze, Multi-hop Manipulation, Ball Tracking). We observe that by optimizing the Modal Transition Loss to address issues including cross-modal hallucination and low utilization efficiency in Interleaving Thinking, our method consistently outperforms various open-source multimodal models on all tasks and achieves the optimal overall performance. We note that for challenging planning reasoning tasks such as Sokoban, the generation module of UMM fails to understand and follow complex reasoning contexts and construct meaningful visual world modeling, and cross-modal hallucination occurs at an extremely high frequency in such tasks. Consequently, the enhancement of the model's reasoning capability via Flow-GRPO in the second training phase is particularly remarkable. Additionally, in the other tasks, the physical world intuition provided by visual modeling plays a more crucial role, and the reasoning paradigm constructed by Interleaving Thinking delivers a far more substantial improvement over the open-source baselines.

Figure~\ref{fig:acc} further demonstrates the synergistic evolution between model performance and cross-modal reward scores throughout the training process. We build two benchmarks adopting the designed VLM-as-Judge reward functions, which quantify the severity of modal isolation by evaluating single-step text and image generation during interleaved thinking. In the figure, $\ell_{\text{und}\to\text{gen}}$ and $\ell_{\text{gen}\to\text{und}}$ respectively represent the reward scores of the corresponding benchmarks.
% 表~\ref{tab:model_performance}展示了前沿多模态大模型以及和Bagel-7B-MoT相似参数量的多模态开源推理模型在4个任务（Sokoban，Maze，Multi-hop Manipulation，Ball Tracking）上的表现。我们观察到，通过优化Modal Transition Loss解决Interleaving Thinking中出现的跨模态幻觉，低利用效能等问题；Ours Method的Performance在所有Task上持续优于各类开源多模态模型，并且在Overall上达到最优表现。我们注意到，对于Sokoban这类困难规划推理问题，UMM的生成模块难以理解并遵循复杂的推理上下文，并给出具有利用意义的视觉世界建模，在该类问题中跨模态幻觉现象尤其高频发生，因此通过Flow-GRPO在第二训练阶段对于模型推理能力的提升尤为显著。此外，在其余任务中，视觉建模提供的物理世界直觉更为重要，Interleaving Thinking构成的推理范式相较于开源基线，提升更加巨大。

% \begin{table}[t]
% \centering
% \small
% \caption{Comparison before and after Flow-GRPO.}
% \label{tab:conflict}
% \setlength{\tabcolsep}{3mm}
% \begin{tabular}{lcccc}
% \toprule
%  & \multicolumn{2}{c}{$\mathcal{R}_{g2u}$} & \multicolumn{2}{c}{$\mathcal{R}_{u2g}$} \\
% \cmidrule(lr){2-3} \cmidrule(lr){4-5}
% Task & Pre & Post & Pre & Post \\
% \midrule
% Sokoban      & 52.80 & 51.20 & 56.00 & 60.80 \\
% Maze         & 64.00 & 65.33 & 80.00 & 88.00 \\
% Manipulation & 82.67 & 81.33 & 49.33 & 58.67 \\
% Ball         & 62.67 & 60.00 & 73.33 & 76.00 \\
% \midrule
% Average      & 65.53 & 64.47 & 64.66 & 70.87 \\
% \bottomrule
% \end{tabular}
% \vspace{-4mm}
% \end{table}

\begin{wrapfigure}{r}{0.45\linewidth}   % 靠右，宽度占当前行宽的45%
    \centering
    \vspace{-3mm}
    \includegraphics[width=\linewidth]{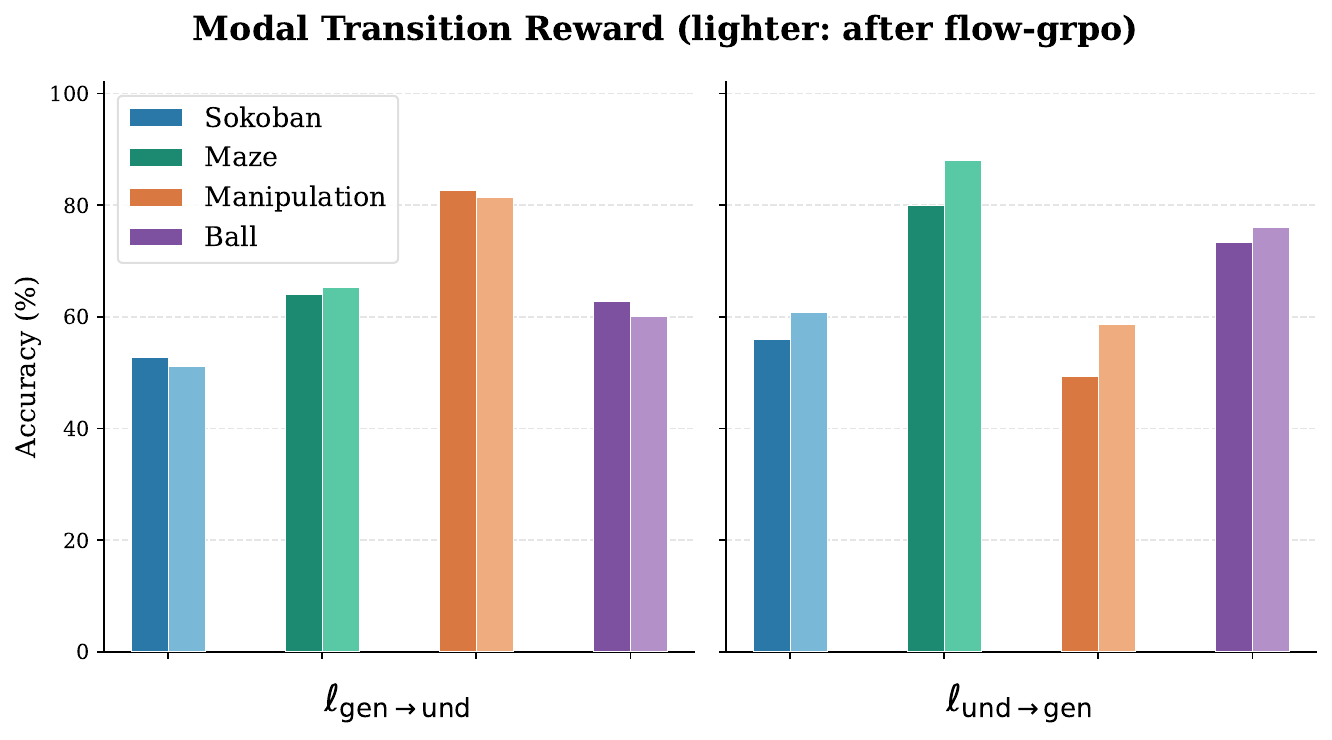}
    \caption{Changes of $\mathcal{R}_{g2u}$ and $\mathcal{R}_{u2g}$ before and after Flow-GRPO training.}
    \label{fig:reflexion}
    \vspace{-1mm}
\end{wrapfigure}

% \newpage
\subsection{Will Reflective SFT and Flow-GRPO Damage Each Other?}
Since the two training phases of Reflective SFT and Flow-GRPO optimize the modal transition loss in different dimensions, we systematically analyze whether multi-objective conflict and catastrophic forgetting occur during this process. Given that prior work has indicated that optimizing UMM with Flow-GRPO would impair the model's reasoning ability, Figure~\ref{fig:acc2} analyzes the model's cross-modal performance on different tasks. The results show that Flow-GRPO can significantly enhance the model's understanding-to-generation process, while keeping the degradation of generation-to-understanding ability within an acceptable range, and even achieves a slight improvement on the Maze task. Detailed per-task numbers are provided in Appendix~\ref{app:conflict} (Table~\ref{tab:conflict}). This demonstrates that our designed two-stage training pipeline can stably and effectively improve the interleaving thinking capability of UMM.

% 由于Reflective SFT和Flow-GRPO两个训练阶段针对不同纬度的模态转换损失进行优化，我们系统性的分析了在该过程中，是否存在多目标拉扯以及灾难性遗忘。因为有先前工作表明，对UMM的flow-grpo优化会损害模型的推理能力，图~\ref{fig:acc2}通过分析不同任务上模型在跨模态过程中的表现，Flow-GRPO可以在显著提升模型从理解到生成过程的同时，将从生成到理解的能力退化控制在可接受范围内，甚至在Maze任务上略有提升。这表明了我们所设计的两阶段训练流程可以稳定有效的提升UMM的Interleaving Thinking推理能力

%%%%%%%%%%%%%%%%%%%%%%%%%%%%%%%%%%%%%%%%
\section{Discussion}
In this paper, we identify an inherent failure mode of interleaved thinking in unified multimodal models. By decomposing it into atomic operations, we pinpoint the phenomenon of modal isolation, wherein the textual and visual modalities alternate in form but fail to transfer information across boundaries. We quantitatively evaluate its manifestation using a carefully designed rubrics-based VLM judge. We reveal that effective interleaved reasoning relies on structural supervision at modal boundaries. More broadly, we provides a principled optimization objective for multimodal reasoning that extends beyond this specific paradigm. We hope this work provides both a diagnostic framework and a practical toolkit for advancing multimodal reasoning in unified models.
% 在该文章中，我们发现了UMM所具备的Interleaving Thinking这一新兴推理范式具有其固有失效模式。我们通过原子化操作对Interleaving Thinking进行拆解，锚定Modal Isolation现象，并使用合理设计的rubrics llm judge量化其表现。我们形式化的定义了这一问题，并揭示了：有效的交错思维并非规模化或监督微调的自然产物，而依赖于模态边界上的结构性监督。更广泛地看，模态转换损失为多模态推理提供了超越该具体范式的原则性优化目标。我们开源了相关模型检查点，训练代码以及高质量图文交错数据集，希望该工作可以启发后续工作推动UMM以及多模态推理的发展，从测试性的合成数据任务扩展到更广泛的领域。

%%%%%%%%%%%%%%%%%%%%%%%%%%%%%%

% \section*{Acknowledgments}
% \label{sec:ack}
% We thank xxx for their contributions and support for the project.

\clearpage
\newpage
\bibliographystyle{plainnat}
\setcitestyle{numbers}
\bibliography{ref}

\clearpage
\newpage
\beginappendix

\section{Training Hyperparameter}
We adopt Bagel-7B-MoT as our base model. All experiments, including Reflective SFT and Flow-GRPO, are conducted on 8×H200 GPUs. Specifically, the code for Reflective SFT is adapted from the official Bagel repository, and the Flow-GRPO code is sourced from its corresponding public code repository. We use Qwen3.5-27B as the judge model, which is deployed on two H200 GPUs via SGLang. In this section, we present the detailed hyperparameter configurations.
% 我们选择Bagel-7B-MoT作为我们的Base Model，所有实验（包括Rflective SFT以及Flow-GRPO）均在 8xH200 上进行. 其中Reflective SFT代码改编自Bagel官方仓库，Flow-GRPO改编自flow grpo代码仓库。我们使用qwen3.5-27b作为judge，该模型使用SGLang部署在2张H200上
% 在该章节中我们展示详细的超参数设置

% 表格 1：Reflective SFT 超参数
\begin{table}[h!]
    \centering
    \caption{Reflective SFT experiment hyperparameters.}
    \label{tab:reflective_sft}
    \small
   \setlength{\tabcolsep}{7mm}
    \begin{tabular}{lc}
        \toprule
        \textbf{Parameter} & \textbf{Value} \\
        \midrule
        expected\_num\_tokens & 34560 \\
        max\_num\_tokens & 34560 \\
        max\_num\_tokens\_per\_sample & 17280 \\
        prefer\_buffer\_before & 17280 \\
        total\_steps & 3020 \\
        warmup\_steps & 150 \\
        save\_every & 300 \\
        lr\_scheduler & cosine \\
        mse\_weight & 10 \\
        ce\_weight & 1 \\
        \bottomrule
    \end{tabular}
\end{table}

% 表格 2：Flow GRPO 超参数
\begin{table}[h!]
   \small
   \setlength{\tabcolsep}{7mm}
    \centering
    \caption{Flow GRPO experiment hyperparameters.}
    \label{tab:flow_grpo}
    \begin{tabular}{lc}
        \toprule
        \textbf{Parameter} & \textbf{Value} \\
        \midrule
        judge\_model & qwen3.5\_27b \\
        train\_batch\_size & 6 \\
        num\_image\_per\_prompt & 16 \\
        num\_batches\_per\_epoch & 1 \\
        learning\_rate & 5e-6 \\
        beta & 0.04 \\
        sde\_window\_size & 3 \\
        seed & 42 \\
        noise\_level & 1.3 \\
        \bottomrule
    \end{tabular}
\end{table}

\section{Analysis of Multi-Objective Conflict}
\label{app:conflict}

Table~\ref{tab:conflict} reports the per-task modality transition rates before and after Flow-GRPO training. ``Pre'' denotes the model after Reflective SFT only; ``Post'' denotes the model after both Reflective SFT and Flow-GRPO. $\mathcal{R}_{g2u}$ measures generation-to-understanding fidelity (image$\to$text), and $\mathcal{R}_{u2g}$ measures understanding-to-generation fidelity (text$\to$image).

\begin{table}[ht]
\centering
\small
\caption{Comparison before and after Flow-GRPO.}
\label{tab:conflict}
\setlength{\tabcolsep}{3mm}
\begin{tabular}{lcccc}
\toprule
 & \multicolumn{2}{c}{$\mathcal{R}_{g2u}$} & \multicolumn{2}{c}{$\mathcal{R}_{u2g}$} \\
\cmidrule(lr){2-3} \cmidrule(lr){4-5}
Task & Pre & Post & Pre & Post \\
\midrule
Sokoban      & 52.80 & 51.20 & 56.00 & 60.80 \\
Maze         & 64.00 & 65.33 & 80.00 & 88.00 \\
Manipulation & 82.67 & 81.33 & 49.33 & 58.67 \\
Ball         & 62.67 & 60.00 & 73.33 & 76.00 \\
\midrule
Average      & 65.53 & 64.47 & 64.66 & 70.87 \\
\bottomrule
\end{tabular}
% \vspace{-4mm}
\end{table}

Several observations emerge from the detailed breakdown: \paragraph{$\mathcal{R}_{u2g}$ improves uniformly.} All four tasks show positive gains after Flow-GRPO, with Manipulation exhibiting the largest improvement (+9.34). This is expected, as Flow-GRPO directly optimizes image generation fidelity to textual instructions. 

\paragraph{$\mathcal{R}_{g2u}$ degradation is bounded.} The largest drop occurs on Ball Tracking (--2.67), while Maze actually improves (+1.33). We hypothesize that improved image quality indirectly benefits subsequent textual observation: when generated images are more faithful, the model's ``observation'' step encounters less distributional shift, partially compensating for any forgetting of the Reflective SFT. 

\paragraph{Net effect is strongly positive.} The average $\mathcal{R}_{u2g}$ gain (+6.21) far exceeds the average $\mathcal{R}_{g2u}$ loss (--1.06), indicating that the two training stages are largely complementary. This supports our design choice of sequential optimization rather than joint multi-task training, which would require careful loss balancing.

\section{Prompt Design}
\label{sec:append_prompt}

In this section, we present the prompt templates designed for our pipeline. These include: 
\begin{itemize}
    \item (1)~prompts for collecting Naive Interleaving Thinking SFT data, which specify the atomic operation cycle (Reason $\to$ Instruct $\to$ Draw $\to$ Observe) and format constraints tailored to each task domain;
    \item (2)~prompts for collecting Reflexion-augmented SFT data, where corrupted images are injected and the model is instructed to detect discrepancies and recover;
    \item (3)~prompts for the VLM-as-Judge reward function used to quantify modality transition loss ($\mathcal{R}_{u2g}$ and $\mathcal{R}_{g2u}$).
\end{itemize}
All prompts maintain a consistent meta-structure across tasks while adapting domain-specific evaluation criteria.
% 在这一节中我们将展示收集Interleaving Thinking SFT数据时所设计的提示词模版：包含Naive SFT Data以及SFT Data with Reflexion。同时我们还将展示用于衡量模态转换损失的基于vlm-as-judge的奖励函数提示词模版

\begin{tcolorbox}[
  colback=iris,   % soft yellow background
  colframe=white,  % warm golden frame
  title=\textbf{Naive SFT Data Collection Prompt Template},
  fonttitle=\bfseries,
  sharp corners
]

**System Prompt:** You will receive a visual reasoning problem based on the \{task name\} environment, including:\\
\\
- A textual question\\
- The original image\\
- A set of raw reasoning steps\\
- The correct answer\\
- A set of intermediate auxiliary images generated during the process of solving the visual reasoning problem\\
\\
Your task is to combine the intermediate auxiliary images to generate an easy-to-understand, logically complete image-text interleaved reasoning process. Your reader will not have access to the intermediate materials (e.g., answer, raw reasoning steps, auxiliary images from the intermediate process). Your explanation will be used as a direct output to users, so it must read naturally and independently.\\
\\
Your output must follow this structure: place your reasoning process and answer between the special tokens \verb|[ICOT_START]| and \verb|[ICOT_END]|.\\
\\
Example:\\
\verb|[ICOT_START]|\\
reasoning process here $\langle$image$\rangle$ (...and so on...) reasoning process here $\langle$image$\rangle$ answer here\\
\verb|[ICOT_END]|\\
\\
**Guidelines:** \\
We model the chain of thought with interleaved text and images as an explicit "Observe and Reflect → Think and Decide → Plan to Draw → Draw" loop. In each reasoning process, first observe the image generated in the previous step, capture and analyze its visual elements in detail (or analyze the input query image at the first step). Then, based on the historical reasoning information and the visual feedback from drawing, provide a detailed reasoning process. Finally, using concise language, naturally derive an accurate and executable drawing plan based on the reasoning content. Please output a scontinuous, natural thinking narrative without explicit headings, bullet points, segmented labels, or rigid structural divisions. Integrate all stages naturally like human reasoning, rather than splitting them into separate labeled parts.\\
\\
**Attention:**\\
- **Reasoning Style:** Simple restatement of the provided analysis is not allowed. Follow the long reasoning style of DeepSeek-R1. Elaborate every step of reasoning in full detail, do not skip intermediate derivations, and generate as much comprehensive reasoning content as possible.\\
- **Strict Image Mapping:** You must include every provided intermediate auxiliary image exactly once, in their original order. Use the \verb|`<image>`| placeholder to insert them. Do not skip, reorder, or hallucinate images.\\
- **No Spoilers:** Do not reveal the final solution or future steps prematurely. The reasoning must appear to unfold step-by-step. Your response must not imply that you possess prior knowledge of the answer or the raw reasoning steps.\\
- **Final Synthesis:** After the last image, provide a summary. And put the final answer in the Latex symbol \verb|`\\boxed{}`|. \\
\\
**User Prompt:**

\end{tcolorbox}

\begin{tcolorbox}[
  colback=iris,   % soft yellow background
  colframe=white,  % warm golden frame
  title=\textbf{Reflective SFT Data Collection Prompt Template},
  fonttitle=\bfseries,
  sharp corners
]
You are helping create augmented training data for a multimodal reasoning model. 
This model performs interleaved text-image reasoning: it generates a text segment 
containing a drawing plan, then generates an image based on that plan, then continues 
with more text, and so on.\\
Below is a multimodal problem and a partial reasoning chain. To simulate an image 
generation error for training purposes, one image has been deliberately replaced 
with a WRONG image.\\

\verb|---|\\
\#\# Original Multimodal Problem\\
\verb|<img>|\\
\verb|{{question_text}}|\\
\#\# Completed Reasoning Steps\\
**Step 1:**\\
\verb|{{step_1_text}}|\\
\verb|<img>|\\
**Step 2:**\\
\verb|{{step_2_text}}|\\
\verb|<img>|\\
... (repeat to step j-1)\\
**Step \verb|{{j}}|:**\\
\verb|{{step_j_text}}|\\
\\
\#\# SIMULATED IMAGE GENERATION ERROR AT STEP \verb|{{j}}|\\
The model attempted to generate an image based on the drawing plan in Step \verb|{{j}}|, 
but the generation went wrong. Here is the WRONG image that was actually produced:\\
\verb|<img>|\\
\#\# Original Text That Followed Step \verb|{{j}}| (for reference)\\
\verb|{{original_next_text}}|\\
\verb|---|\\

\#\# Your Task
The wrong image shown above was generated instead of the correct one. 
You need to rewrite the text that follows this wrong image (replacing the 
original text shown above for reference).\\
\\
Your rewritten text must follow this three-part structure:\\
1. **Observe and describe the error**: Carefully examine the WRONG image. 
   Describe what you actually see in it and explain specifically why it does 
   NOT match what was expected based on Step \verb|{{j}}|'s reasoning and drawing plan.\\
2. **Continue reasoning**: Acknowledge that an image generation error occurred, 
   but decide to move forward rather than redraw. Based on what the CORRECT 
   image SHOULD have shown (as intended by Step \verb|{{j}}|), continue the logical 
   reasoning for the next step.\\
3. **Drawing plan for the next step**: Provide a clear, executable drawing 
   instruction for the NEXT image to generate. This drawing plan should be 
   consistent with the original next step's intended target.\\
\\
Write in first person. Use natural hesitation markers like "Hmm...", 
"Wait,", "Let me look more closely...", or "Actually..." where appropriate.\\
\\
Please think carefully about the situation first. Analyze what the correct state 
should be vs. what the wrong image shows. Then provide your output in the JSON blob.
\end{tcolorbox}

\begin{tcolorbox}[
  colback=iris,   % soft yellow background
  colframe=chisa,  % warm golden frame
  title=\textbf{$\mathcal{R}_{u2g}$},
  fonttitle=\bfseries,
  sharp corners
]
\#\# Role \& Identity\\
You are a strict visual verifier for the image generated in interleaved text-image reasoning content.\\
\#\# Task Background\\
Unified Generation and Understanding Models (UMM) natively integrate autoregressive text generation capability and flow-matching based image generation capability. For complex multimodal problem-solving (e.g., geometry reasoning), these models adopt an interleaved text-image multi-modal chain-of-thought (CoT) reasoning paradigm: they alternate between generating textual reasoning steps and auxiliary images in a cyclic reasoning process, to combine abstract symbolic reasoning from text and physical world intuition from images. This reasoning paradigm is defined as a generic "think-decide -> drawing plan -> drawing -> observe-reflect" loop, wherein the textual reasoning generated in each round first observes and reflects upon the visual elements in the image produced in the previous step (if any) to assist subsequent reasoning and decision-making, and finally formulating a detailed drawing plan for the next step.\\
The model receives an original multimodal problem and a completed text-image reasoning context. It must generate the next auxiliary image. The generated image is compared with a ground-truth auxiliary image. Your job is to produce one binary correctness judgment: whether the generated image is logically equivalent to the ground-truth image for this reasoning step.\\
\\
\#\# Core Evaluation Task\\
You will be given:\\
1. The original multimodal problem image and problem text.\\
2. The full context of completed interleaved multi-modal CoT reasoning steps, where the corresponding images at their respective positions are represented by the `<auxiliary-image>` placeholder.\\
3. The image to be evaluated, which is generated following the explicit visualization instructions of the previous step's textual reasoning.\\
4. The ground-truth image.\\
\\
\#\# Strict Evaluation Principles\\
Focus only on correctness, not standalone aesthetic preference. The generated image does not need to be pixel-identical to the ground truth, but it must be logically equivalent for the current reasoning step.\\
Score = 1 only if all of the following are true:\\
- The generated image and the ground-truth image represent the same task state, same relevant objects, same spatial relations, same markings/annotations, and same action/result implied by the current step.\\
- The generated image faithfully reflects the core intent of the immediately preceding textual drawing instruction in the reasoning context.\\
- Non-target regions, objects, labels, or structural elements that should remain unchanged are logically preserved.\\
- Any visual differences are only superficial style-level differences and do not alter the task-relevant meaning.\\
- The image is clear enough that all task-relevant entities and relations can be reliably judged.\\
\\
Score = 0 if any task-relevant object, relation, state, annotation, action, or unchanged region differs from the ground truth. Also score 0 if the image is too blurry, distorted, incomplete, or artifact-corrupted to verify logical equivalence.\\
Use the original problem and reasoning context only to understand what aspects are task-relevant. The final decision must be based on logical equivalence between the generated image and the ground-truth image.
\end{tcolorbox}

\begin{tcolorbox}[
  colback=iris,   % soft yellow background
  colframe=chisa,  % warm golden frame
  title=\textbf{$\mathcal{R}_{g2u}$},
  fonttitle=\bfseries,
  sharp corners
]
\#\# Role \& Identity\\
You are a strict text verifier for the text generated in interleaved text-image reasoning content.\\
\#\# Task Background\\
Unified Generation and Understanding Models (UMM) natively integrate autoregressive text generation capability and flow-matching based image generation capability. For complex multimodal problem-solving (e.g., geometry reasoning), these models adopt an interleaved text-image multi-modal chain-of-thought (CoT) reasoning paradigm: they alternate between generating textual reasoning steps and auxiliary images in a cyclic reasoning process, to combine abstract symbolic reasoning from text and physical world intuition from images. This reasoning paradigm is defined as a generic "think-decide -> drawing plan -> drawing -> observe-reflect" loop, wherein the textual reasoning generated in each round first observes and reflects upon the visual elements in the image produced in the previous step (if any) to assist subsequent reasoning and decision-making, and finally formulating a detailed drawing plan for the next step.\\
\\
The model receives an original multimodal problem and a completed text-image reasoning context. It must generate the next reasoning text step. Your job is to produce one binary correctness judgment: whether the generated text correctly fulfills the requirements of this reasoning step.\\
\\
\#\# Core Evaluation Task\\
You will be given:\\
1. The original multimodal problem image and problem text.\\
2. The full context of completed interleaved multi-modal CoT reasoning steps, where the corresponding images at their respective positions are represented by the `<auxiliary-image>` placeholder.\\
3. The image generated in the immediately preceding step (if any), which the model must observe and reflect upon. If there is no preceding image (i.e., this is the very first reasoning step), you will only see the problem image and must judge whether the model correctly observes the visual elements in the problem image instead.\\
4. The generated text to be evaluated.\\
5. The step type: "non-final" (a \verb|<think>...</think>| block ending with \verb|<create_image>|) or "final" (an \verb|<ans>...</ans>| block that concludes the entire reasoning chain).\\
\\
This is an intermediate reasoning step. The generated text must follow the format \verb|<think>...</think>| and end with \verb|<create_image>|. Score = 1 only if ALL of the following are true:\\
- The generated text is wrapped in \verb|<think>...</think>| and ends with \verb|<create_image>|.\\
- The generated text correctly observes and describes the task-relevant visual elements from the immediately preceding image (or from the problem image, if this is the first step with no preceding image). The observations must be accurate and grounded in what is actually visible.\\
- The reasoning is logically coherent, mathematically or spatially correct, and consistent with the problem rules and the preceding context.\\
- The final part of the think block gives a clear, specific, and executable visual drawing/diagram instruction that faithfully follows from the reasoning.\\
\\
Score = 0 if the format is wrong, the visual observations are hallucinated or inaccurate, the reasoning is flawed or contradictory, the drawing instruction is missing or unreasonable, or the text is otherwise inadequate for guiding the next image generation step.\\
Use the original problem and reasoning context only to understand what aspects are task-relevant. You must strictly follow the required JSON output format.
\end{tcolorbox}

\section{Task Examples and Model Responses}
We synthesize all Interleaving Thinking reasoning chain data using Gemini3.1-Flash-Lite. For raw data generation (including questions and intermediate images), the code for Sokoban and Maze is adapted from Game-RL, while Ball Tracking and Multi-hop Manipulation are fully original implementations. In this section, we present all 4 tasks' examples and examples of Interleaved Thinking reasoning chains generated by the trained models.
% 对于所有Interleaving Thinking 思维链数据，我们均采用Gemini3.1-Flash-Lite合成思维链数据。对于原始数据（包括问题以及中间图像）的合成，Sokoban和Maze代码改编自Game-RL，Ball Tracking和Multi-hop Manipulation均为原创
% 在这一节中我们将展示多个任务示例以及训练后模型输出的Interleaving Thinking思维链示例

\begin{table}[htbp]
\centering
\caption{Detailed Dataset Information}
\label{tab:datas}
\small
\begin{tabular}{l c c} % 改为普通的 tabular，去掉宽度限制和列间距拉伸
\toprule
\textbf{Task} & \textbf{Size-RL} & \textbf{Size-SFT} \\
\midrule
Sokoban      & 6074           & 15994 \\
Maze         & 3945           & 14730 \\
Manipulation & 4034           & 16000   \\
Ball Tracking & 4615          & 16000   \\
\bottomrule
\end{tabular}
\end{table}

\begin{tcolorbox}[
  colback=iris,   % soft yellow background
  colframe=white,  % warm golden frame
  title=\textbf{Ball},
  fonttitle=\bfseries,
  sharp corners
]
\textbf{Images}
\begin{center}
\includegraphics[width=0.3\linewidth]{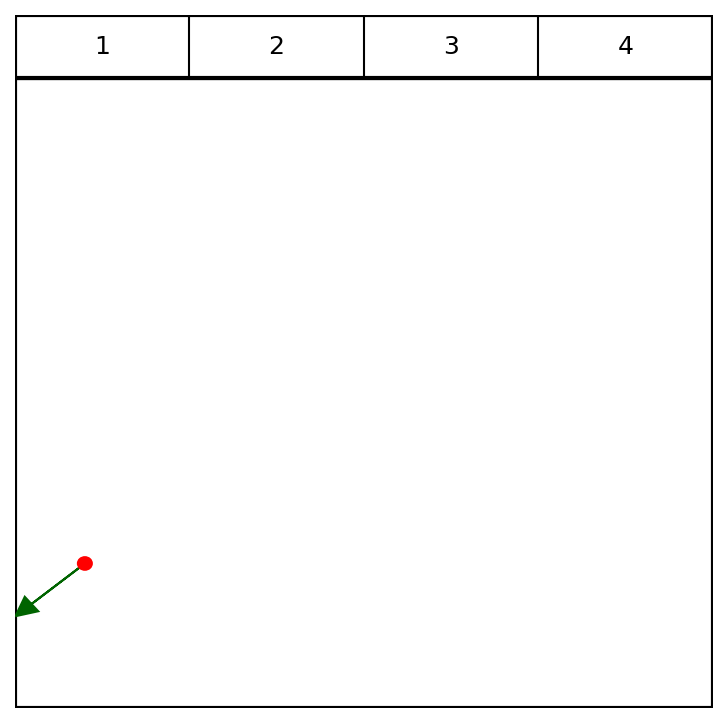}
\end{center}

\textbf{Question}:\\
**Rules:**1. There is a red ball and a green arrow; the arrow indicates the ball's initial direction of motion; 2. At the top of the image, there are several rectangular holes arranged horizontally, numbered from left to right starting at 1; 3. The black boundaries of the image represent solid walls. When the ball hits a wall, it undergoes ideal reflection; 4. The red ball is considered a point mass-its size is negligible; 5. The ball reflects according to the following rules:\\
- The component of velocity perpendicular to the wall is reversed in direction;\\
- The component of velocity parallel to the wall remains unchanged;\\
- The ball moves at a constant speed in an ideal environment until it enters one of the holes.\\
**Question:** Estimate which hole the red ball will enter first during its motion.
\end{tcolorbox}

\begin{tcolorbox}[
  colback=iris,   % soft yellow background
  colframe=white,  % warm golden frame
  title=\textbf{Multi-hop Manipulation},
  fonttitle=\bfseries,
  sharp corners
]
\textbf{Images}
\begin{center}
\includegraphics[width=0.3\linewidth]{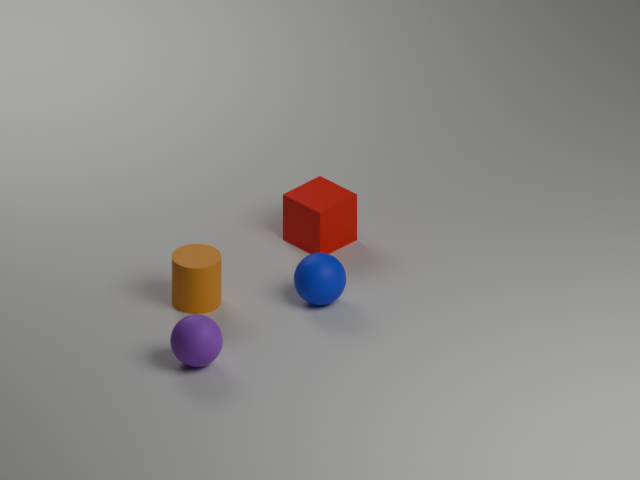}
\end{center}

\textbf{Question}:\\
You are given an initial arrangement of objects viewed from an oblique front perspective. In this view, objects closer to the camera are considered 'front,' and the 'left' and 'right' sides of the scene correspond to the left and right of the frame.Perform the following operations in sequence:\\
1. Change the orange cylinder into a gray cylinder.\\
2. Change the blue sphere into a red cylinder.\\
After completing all operations, what is the object in front of the red cuboid? \\Options:\\
A. red cylinder\\ B. purple sphere\\
C. gray cylinder\\ D. red cuboid
\end{tcolorbox}

% \begin{tcolorbox}[
%   colback=iris,   % soft yellow background
%   colframe=white,  % warm golden frame
%   title=\textbf{Maze},
%   fonttitle=\bfseries,
%   sharp corners
% ]
% \textbf{Images}
% \begin{center}
% \includegraphics[width=0.5\linewidth]{figures/maze.png}
% \end{center}

% \textbf{Question}:\\
% **Rules:**1. This is a maze mini-game.The player needs to navigate around obstacles to reach the destination and achieve victory. 2. The red circle represents the player, the green block is the goal and the blue blocks are obstacles. 3. The player can only move within the white blocks. 4. The coordinates are given in the format (row, col), where row represents the vertical position and col represents the horizontal position. Find the path to the finish and count the number of turns it takes to get there.
% \end{tcolorbox}

\begin{tcolorbox}[
  colback=iris,   % soft yellow background
  colframe=white,  % warm golden frame
  title=\textbf{Sokoban},
  fonttitle=\bfseries,
  sharp corners
]
\textbf{Images}
\begin{center}
\includegraphics[width=0.3\linewidth]{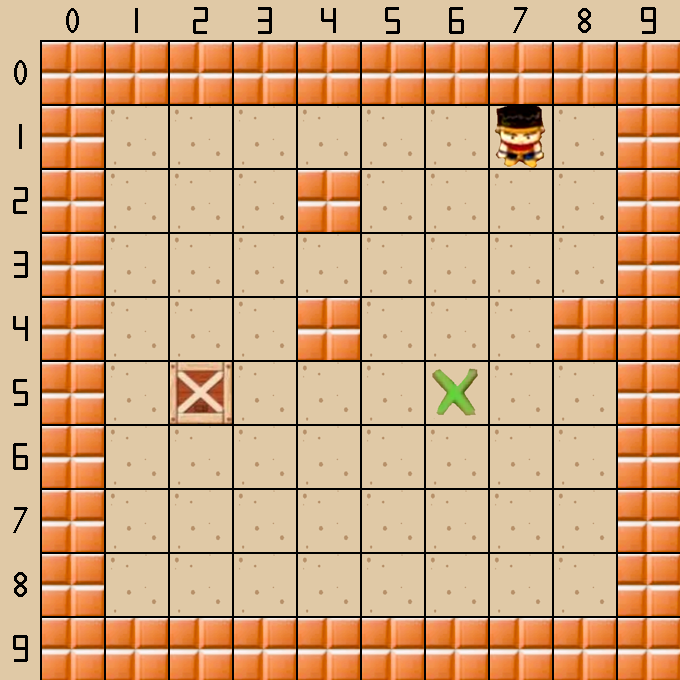}
\end{center}

\textbf{Question}:\\
This is a Sokoban puzzle where cartoon person is player, green X is target, brown box with X is box to push, brown tiles are walls, and light brown areas are movable spaces.The coordinates (x, y) in this puzzle represent the matrix format.What is the minimum number of moves needed to solve this puzzle? Output the option number instead of the num of steps\\
Options:\\\verb|[1]| 15\quad \verb|[2]| 12\\
\verb|[3]| 11\quad \verb|[4]| 14\\
\verb|[5]| 21\quad \verb|[6]| 17\\
\verb|[7]| 13\quad \verb|[8]| 16
\end{tcolorbox}

\begin{tcolorbox}[
  colback=iris,   % soft yellow background
  colframe=chisa,  % warm golden frame
  title=\textbf{Model Responses Example 1},
  fonttitle=\bfseries,
  sharp corners
]
\textbf{Model Response}:\\
\verb|<think>| To begin solving this visual reasoning problem, I first need to examine the initial state of the scene. Looking at the original image, I can identify four distinct objects arranged on a neutral gray surface. Closest to the camera, or \"front,\" is a purple sphere. Directly behind it sits an orange cylinder. To the right of this column, there is a blue sphere, and further behind that blue sphere is a red cuboid. Based on the perspective provided, the purple sphere and blue sphere are on the same horizontal plane relative to the camera, while the orange cylinder and red cuboid are positioned further back.\\
The first instruction is to change the orange cylinder into a gray cylinder. I observe the orange cylinder located in the middle-left area of the arrangement, positioned behind the purple sphere. To execute this change, I will mentally replace the color property of this specific object. I plan to visualize this by rendering the scene with the cylinder at that same position but with a gray matte finish instead of its original orange color. \verb|</think><create_image>|

\begin{center}
\includegraphics[width=0.3\linewidth]{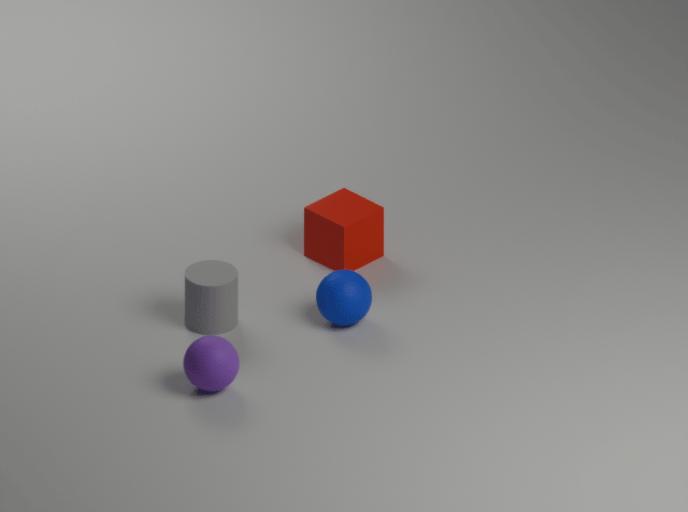}
\end{center}
\verb|<think>| With the first step complete, I observe the updated scene. We now have a gray cylinder where the orange one used to be. The next instruction is to change the blue sphere into a red cylinder. I look across to the right side of the frame where the blue sphere is located. This sphere is positioned directly in front of the red cuboid. My plan is to modify this object's shape and color: the spherical geometry will be replaced by a cylindrical one, and the color will shift from blue to red. \verb|</think><create_image>|

\begin{center}
\includegraphics[width=0.3\linewidth]{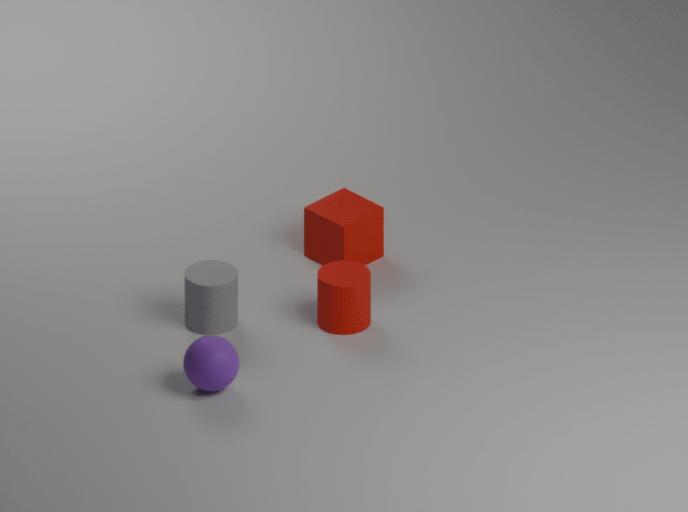}
\end{center}
\verb|<ans>| Now that both operations are finished, I need to analyze the final configuration to answer the question: \"What is the object in front of the red cuboid?\" Looking at the final image, I locate the red cuboid at the back right of the group. Directly in front of it, occupying the position where the blue sphere used to be, is the newly created red cylinder. On the left side of the scene, the purple sphere remains in the front, and the gray cylinder is behind it. Since the red cylinder is the object situated immediately between the red cuboid and the camera's perspective line, it is the object in front of the red cuboid.\\The sequence of transformations changed the orange cylinder to gray and the blue sphere to a red cylinder. In the final state, the red cuboid is the furthest object in its column, and the object positioned directly in front of it is the red cylinder.\\Final Answer: The object in front of the red cuboid is the red cylinder. \verb|\\boxed{A}| \verb|</ans>|

\end{tcolorbox}

\begin{tcolorbox}[
  colback=iris,   % soft yellow background
  colframe=chisa,  % warm golden frame
  title=\textbf{Model Responses Example 2},
  fonttitle=\bfseries,
  sharp corners
]
\textbf{Images}
\begin{center}
\includegraphics[width=0.25\linewidth]{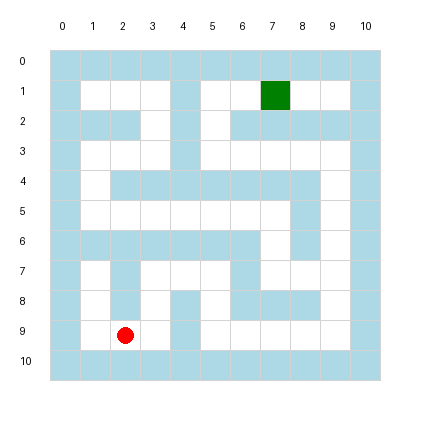}
\end{center}

\textbf{Question}:\\
**Rules:**1. This is a maze mini-game.The player needs to navigate around obstacles to reach the destination and achieve victory. 2. The red circle represents the player, the green block is the goal and the blue blocks are obstacles. 3. The player can only move within the white blocks. 4. The coordinates are given in the format (row, col), where row represents the vertical position and col represents the horizontal position. Find the path to the finish and count the number of turns it takes to get there.
\\
\textbf{Model Response}:\\
\verb|<think>| To solve the maze, I will start by identifying the player's position at (9, 2) and the goal at (1, 7). The goal is to navigate the white paths while avoiding the blue obstacles. Initially, I will test the path to the left. I observe that moving from (9, 2) to (9, 1) and then upward leads to a dead end at (7, 1). \verb|</think><create_image>|

\begin{center}
\includegraphics[width=0.25\linewidth]{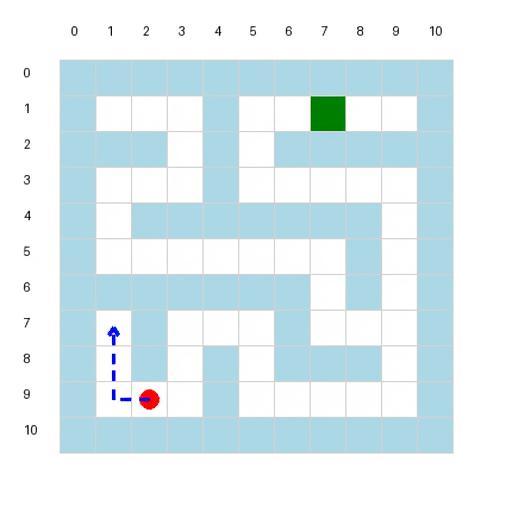}
\end{center}
\verb|<think>| Since the first path is blocked, I will backtrack to the starting point at (9, 2) and explore the path to the right. Moving from (9, 2) to (9, 3), I then head upwards to (7, 3), turn right toward (7, 5), go down to (9, 5), and continue right to (9, 9). From there, I navigate upward along the right edge of the maze to (3, 9). Turning left, I proceed to (3, 5), then move up to (1, 5), and finally head right to reach the goal at (1, 7). By tracing this path and making turns whenever the direction changes, I count the total number of turns required to successfully reach the destination. \verb|</think><create_image>|

\begin{center}
\includegraphics[width=0.25\linewidth]{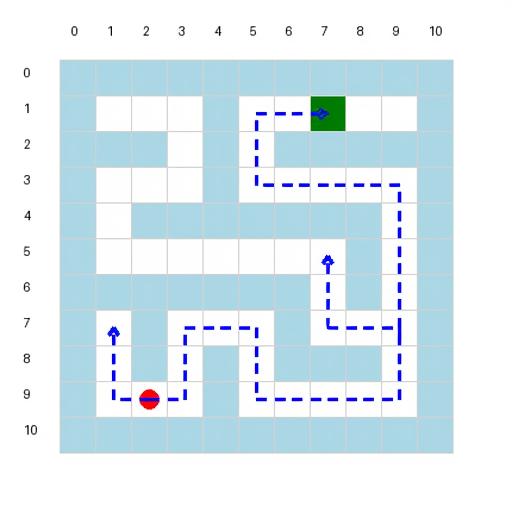}
\end{center}
\verb|<ans>| The path taken involves several directional changes: starting from (9, 2), the turns occur at (9, 3), (7, 3), (7, 5), (9, 5), (9, 9), (3, 9), (3, 5), and (1, 5). Counting these navigation pivots, the total number of turns is 8. \verb|\\boxed{8}| \verb|</ans>|

\end{tcolorbox}

\end{document}